\documentclass[runningheads]{llncs}

\usepackage{eccv}

\usepackage{eccvabbrv}
\usepackage{wrapfig}
\usepackage{graphicx}
\usepackage{booktabs}

\usepackage[accsupp]{axessibility}  
\makeatletter
\renewcommand\section{\@startsection{section}{1}{\z@}%
  {-10\p@ \@plus -2\p@ \@minus -2\p@}%
  {5\p@ \@plus 1\p@ \@minus 1\p@}%
  {\normalfont\large\bfseries\boldmath
   \rightskip=\z@ \@plus 8em\pretolerance=10000}}

\renewcommand\subsection{\@startsection{subsection}{2}{\z@}%
  {-7\p@ \@plus -1.5\p@ \@minus -1\p@}%
  {3\p@ \@plus 1\p@ \@minus 1\p@}%
  {\normalfont\normalsize\bfseries\boldmath}}
\makeatother
\usepackage{hyperref}

\usepackage{orcidlink}

\begin{document}


\title{OnEvoMemory: Evolving Memory through Online Robot Rollouts for Pretrained Robot Policies}

\titlerunning{OnEvoMemory: Evolving Memory for Robot Policies}

\author{
Zhongxi Chen\inst{1}\textsuperscript{*,\textdagger}
\and
Shenqi Zong\inst{2}\textsuperscript{*}
}

\authorrunning{Z. Chen and S. Zong}

\institute{
Shanghai Jiao Tong University
\and
Tsinghua University
}

\maketitle

\renewcommand{\thefootnote}{*}
\footnotetext{Equal contribution.}

\renewcommand{\thefootnote}{\textdagger}
\footnotetext{Corresponding author.}

\begin{abstract}

Long-horizon robot manipulation requires policies to track completed subtasks
and critical interaction events. However, existing memory mechanisms heavily
rely on external models or predefined update rules. To address this, we propose
\textbf{OnEvoMemory}, a value-guided memory module for pretrained robot
policies. It maintains recent context, high-value experiences, and salient
transitions, while learning which experiences should be retained from
trajectory outcomes. Offline demonstrations initialize the memory prior,
whereas successful and unsuccessful online rollouts refine memory selection, helping the policy recognize task-stage transitions and
avoid repeating completed subtasks. Experiments on long-horizon manipulation
benchmarks show that OnEvoMemory improves the performance of the
base VLA policy through both offline initialization and online memory evolution.
\keywords{Long-horizon manipulation \and Vision-language-action models}
\end{abstract}

\section{Introduction}
\label{sec:intro}

Embodied policies, such as vision-language-action (VLA) models~\cite{kim2025openvla} and diffusion policies~\cite{chi2025diffusionpolicy}, continue to face substantial challenges in long-horizon task execution and often benefit from additional memory mechanisms~\cite{koo2026hamlet,shah2026halo,shi2026memoryvla}. Existing approaches commonly augment policies with fixed temporal windows, similarity-based retrieval, or manually designed writing rules~\cite{koo2026hamlet,shah2026halo,yang2026memorywamefficientworldaction,zeng2026kemo,zhu2026weavela}. However, these methods typically predetermine what information should be retained, making it difficult to adapt the focus of memory to the requirements of different tasks based on actual interaction outcomes. In long-horizon manipulation, certain events---such as establishing a successful grasp, losing contact with an object, or entering a new task stage---can have a considerable influence on subsequent decisions~\cite{zeng2026kemo,zhu2026weavela}. This raises a central question: how can a model learn what is worth remembering?

\begin{figure}[t]
    \centering
    \includegraphics[width=\linewidth]{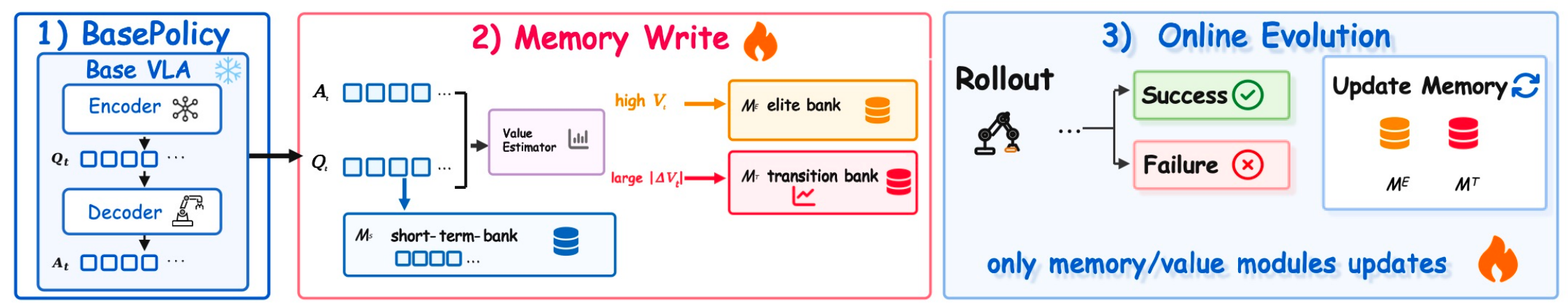}
    \caption{\textbf{Overview of OnEvoMemory.}
    A frozen base VLA produces action-query representations, which are organized
    into short-term, elite, and transition memories through value-guided writing.
    Successful and unsuccessful robot rollouts further update only the value and
    memory modules, enabling online memory evolution without modifying the
    pretrained policy.}
    \label{fig:overview}
\end{figure}

To address this question, as illustrated in Fig.~\ref{fig:overview}, we introduce
\textbf{OnEvoMemory}, a value-guided
memory system that is initialized from offline experience and continuously
evolves through online robot rollouts. 

Our main contributions are summarized as follows:
\begin{itemize}
    \setlength{\itemsep}{1pt}
    \setlength{\parskip}{0pt}
    \setlength{\parsep}{0pt}
    \setlength{\topsep}{2pt}
    \setlength{\partopsep}{0pt}
    \setlength{\leftmargin}{1.5em}

    \item We introduce a hierarchical memory architecture consisting of a
    short-term raw buffer, an elite experience bank, and a transition bank,
    which capture recent context, high-value experiences, and salient value
    transitions, respectively.

    \item We develop a learnable, value-guided memory-writing mechanism that
    identifies task-relevant experiences without relying on fixed temporal
    windows or manually defined writing rules.

    \item We propose an online memory-evolution approach that uses successful
    and unsuccessful rollouts to refine the memory mechanism while keeping the
    underlying policy frozen.
\end{itemize}








\section{Related Work}

Recent memory-augmented robot policies mainly differ in how they select and maintain historical information~\cite{koo2026hamlet,shi2026memoryvla,shah2026halo,
yang2026eventvla}. One line of work relies on external vision--language models to identify task-relevant history. For example, BPP uses an off-the-shelf VLM to detect semantic keyframes, while MemER employs a high-level VLM to select previous observations and produce subtask instructions for a low-level policy \cite{mark2026bpp,sridhar2026memer}. Although these methods benefit from strong semantic priors, their memory selectors are decoupled from the underlying robot policy, making it difficult to continually adapt memory selection from online interaction outcomes.

Another line of work adopts predefined memory structures or update heuristics~\cite{zeng2026kemo,zhu2026weavela}. MemoryWAM, for instance, combines a sliding window of recent observations, task-boundary anchor frames, and compact gist tokens to efficiently preserve long-range context \cite{yang2026memorywamefficientworldaction}. Such designs provide effective and computationally efficient history representations, but what should be retained is largely determined by the initial memory mechanism. In contrast, OnEvoMemory uses rollout outcomes to continually refine a value-guided memory writer, enabling memory selection to evolve online while keeping the pretrained policy frozen.

\section{Method}
\label{sec:method}

\textbf{Overview.}
We propose \textbf{OnEvoMemory}, a value-guided memory mechanism initialized from offline demonstrations and continuously adapted through online robot rollouts. We instantiate it on a token-based VLA policy, while the proposed mechanism can also be applied to policies that expose intermediate action representations.

Given an observation \(o_t=(I_t^{1:V},s_t,\ell)\), where \(I_t^{1:V}\), \(s_t\), and \(\ell\) denote multi-view images, proprioceptive states, and a language instruction, respectively, the base policy produces action-query representations \(Q_t=E_\theta(o_t)\in\mathbb{R}^{H\times d}\), where \(H\) is the action horizon.

OnEvoMemory maintains an episode-local memory \(\mathcal{M}_t=\{\mathcal{M}_t^S,\mathcal{M}_t^E,\mathcal{M}_t^T\}\), consisting of a short-term raw buffer, an elite experience bank, and a transition bank. The current action queries retrieve historical representations from these components to form \(C_t=\operatorname{Retrieve}(Q_t,\mathcal{M}_t)\), which is incorporated through gated cross-attention as \(\widetilde{Q}_t=Q_t+g_t\operatorname{MHA}(Q_t,C_t,C_t)\). The base action decoder then predicts an action chunk \(\hat{A}_t=D_\theta(\widetilde{Q}_t)\). Memory is always read before the current experience is written, preventing the current state from being immediately retrieved within the same decision step.

\subsection{Value-Guided Hierarchical Memory}

OnEvoMemory represents interaction history using three complementary memory
components. The short-term buffer \(\mathcal{M}^S\) stores recent action-query
representations in a fixed-capacity FIFO queue, preserving local visual,
linguistic, and proprioceptive context. The elite experience bank
\(\mathcal{M}^E\) retains high-value state--action experiences, such as stable
grasps and successfully completed intermediate stages, while the transition
bank \(\mathcal{M}^T\) preserves salient changes in trajectory quality,
including task progress, contact loss, failure, and recovery events.

Memory writing is guided by an action-conditioned value estimator
\((V_t,k_t,v_t)=F_\phi(\widetilde{Q}_t,A_t)\), where \(V_t\) provides a
trajectory-outcome-aligned writing score, and \(k_t\) and \(v_t\) denote the
retrieval key and stored representation, respectively. High-value experiences
are written to \(\mathcal{M}^E\), whereas experiences with large temporal value
changes \(\lvert V_t-V_{t-1}\rvert\) are written to \(\mathcal{M}^T\). This
enables the system to identify useful experiences and salient transitions
without relying on fixed temporal windows or manually specified event rules.

During retrieval, each long-term bank returns both semantically relevant and
recently written experiences, following
\(\mathcal{R}_t^b=\operatorname{TopK}_{\mathrm{sim}}(Q_t,\mathcal{M}_t^b)
\cup\operatorname{RecentK}(\mathcal{M}_t^b)\), where \(b\in\{E,T\}\).
The retrieved representations are combined with the short-term buffer to form
the memory context \(C_t\).

\subsection{Offline Initialization and Online Evolution}

During offline initialization, expert demonstrations are replayed in temporal
order, such that earlier experiences populate the memory banks and subsequent
observations retrieve the accumulated memory for action prediction. Action
supervision and trajectory outcomes jointly initialize the memory prior and
train the memory reading, writing, and injection modules.

During online rollouts, both successful and unsuccessful trajectories provide
new outcome supervision, enabling the value estimator and memory writer to
continually revise which experiences should be retained and reused. The
pretrained vision--language backbone and action policy remain frozen throughout
online adaptation, while only the value estimator and memory-related modules
are updated. OnEvoMemory therefore evolves its memory selection and utilization
while preserving the manipulation capability of the pretrained policy.

\section{Experiments}
\label{sec}

\textbf{Experimental setup.}
We instantiate OnEvoMemory on QwenOFT~\cite{community2026starvla,kim2025finetuning} as the base VLA policy and evaluate it
on the ten tasks of LiberoLong-10~\cite{liu2023libero} and two long-horizon tasks from RMBench~\cite{chen2026rmbench}:
SwapBlocks and SwapT. We compare three settings: the original base policy,
the policy augmented with offline-initialized memory, and the same system
after one round of online memory evolution. During online adaptation, we
collect 20 rollout trajectories per task on LiberoLong-10 and 10 trajectories
per task on RMBench. The pretrained QwenOFT policy remains frozen, and only
the value estimator and memory-related modules are updated.

\begin{wraptable}{r}{0.58\textwidth}
\vspace{-1.0\baselineskip}
\centering
\small
\setlength{\tabcolsep}{3pt}
\caption{Success rates (\%). Offline initializes memory from demonstrations,
while Online further updates it with one round of robot rollouts.}
\label{tab:main_results}

\resizebox{\linewidth}{!}{
\begin{tabular}{lccc}
\toprule
Benchmark / Task & Base VLA & + Offline & + Online \\
\midrule
LiberoLong-10 Avg.  & 86.2 & 88.6 & \textbf{90.2} \\
RMBench SwapBlocks & 0    & 10.0   & \textbf{14.0}   \\
RMBench SwapT      & 0    & 8.0    & \textbf{10.0}   \\
\bottomrule
\end{tabular}
}
\vspace{-0.8\baselineskip}
\end{wraptable}
\textbf{Results and analysis.}
As shown in Table~\ref{tab:main_results}, offline memory initialization improves
the average success rate on LiberoLong-10 from 86.2\% to 88.6\%, and one round
of online memory evolution further raises it to 90.2\%. On the more challenging
RMBench tasks, the base policy achieves 0\% success on both SwapBlocks and
SwapT. Offline memory increases the success rates to 10\% and 8\%, respectively,
which are further improved to 14\% and 10\% after online adaptation.

Qualitatively, memory is particularly helpful around task-stage transitions.
By retaining high-value experiences associated with completed subtasks, the
policy is less likely to lose track of its current progress and repeat actions
corresponding to stages that have already been completed. The additional gains
from online adaptation may be attributed to the unsuccessful rollouts collected
during interaction: these negative examples provide corrective supervision for
both the value estimator and the memory-writing mechanism, helping the system
revise overestimated experiences and retain events that are more informative
for subsequent decisions. While the improvements remain moderate and the
RMBench success rates are still limited, the results suggest that offline memory
provides a useful initialization and that rollout-based updates can further
refine memory selection without modifying the pretrained policy.
\section{Conclusion}

We studied how pretrained robot policies can learn what to remember for
long-horizon manipulation. OnEvoMemory combines offline memory initialization
with rollout-driven online refinement, helping the policy preserve task
progress and avoid repeating completed subtasks. Experiments suggest that
memory selection can be improved from both successful and unsuccessful
experience while keeping the base policy frozen.

\bibliographystyle{splncs04}
\bibliography{main}
\end{document}